\documentclass{article}

\PassOptionsToPackage{numbers}{natbib}

\usepackage[preprint]{neurips_2020}

\usepackage[utf8]{inputenc} % allow utf-8 input
\usepackage[T1]{fontenc}    % use 8-bit T1 fonts
\usepackage{hyperref}       % hyperlinks
\usepackage{url}            % simple URL typesetting
\usepackage{booktabs}       % professional-quality tables
\usepackage{amsfonts}       % blackboard math symbols
\usepackage{nicefrac}       % compact symbols for 1/2, etc.
\usepackage{microtype}      % microtypography
\usepackage{graphicx}
\usepackage{subfigure}
\usepackage{multirow}
\usepackage{amsmath}
\usepackage{enumitem}
\usepackage[capitalize]{cleveref}
\usepackage{caption}
\usepackage{tablefootnote}
\usepackage{paralist}

\title{Real-time Tropical Cyclone Intensity Estimation by Handling Temporally Heterogeneous Satellite Data}

\author{%
  Boyo Chen \\
  Center for Weather Climate \\
  and Disaster Research \\
  National Taiwan University \\
  Taipei, Taiwan \\
  \texttt{boyochen722@gmail.com} \\
  \And
  Buo-Fu Chen \\
  Center for Weather Climate \\
  and Disaster Research \\
  National Taiwan University \\
  Taipei, Taiwan \\
  \texttt{bfchen777@gmail.com} \\
  \And
  Yun-Nung Chen \\
  Department of Computer Science \\
  National Taiwan University \\
  Taipei, Taiwan \\
  \texttt{y.v.chen@ieee.org} \\
}

\begin{document}

\maketitle

\begin{abstract}

Analyzing big geophysical observational data collected by multiple advanced sensors on various satellite platforms promotes our understanding of the geophysical system. For instance, convolutional neural networks (CNN) have achieved great success in estimating tropical cyclone (TC) intensity based on satellite data with fixed temporal frequency (e.g., ~3 h). 
However, to achieve more timely (under 30 min) and accurate TC intensity estimates, a deep learning model is demanded to handle temporally-heterogeneous satellite observations. 
Specifically, infrared (IR1) and water vapor (WV) images are available under every 15 minutes, while passive microwave rain rate (PMW) is available for about every 3 hours. Meanwhile, the visible (VIS) channel is severely affected by noise and sunlight intensity, making it difficult to be utilized.
Therefore, we propose a novel framework that combines generative adversarial network (GAN) with CNN. The model utilizes all data, including VIS and PMW information, during the training phase and eventually uses only the high-frequent IR1 and WV data for providing intensity estimates during the predicting phase. 
Experimental results demonstrate that the hybrid GAN-CNN framework achieves comparable precision to the state-of-the-art models, while possessing the capability of increasing the maximum estimation frequency from 3 hours to less than 15 minutes.
\emph{Github link for the paper will be provided.}

\end{abstract}

\section{Introduction}
\label{sec:intro}

Tropical cyclone (TC) is a type of low-pressure weather systems form and develop on the warm tropical ocean.
It is characterized by intense rotating winds and severe rainfall associated with eyewall clouds and spiral rainbands.
A TC hitting the land poses severe threats to society by producing gusty wind, sea surge, flooding, and landslide.

TC intensity (i.e., the maximum sustained surface wind near the center) is one of the most critical factors in disaster management. The state-of-the-art SATCON \cite{velden2014update}, widely used in operational forecasting, estimates TC intensity based on consensus decision-making procedures using infrared images from geostationary satellites and other observation from low-Earth-orbit satellites. Note that high quality SATCON estimates could be obtained with an approximate three-hour frequency.

\begin{figure}[t]
\centering
    \includegraphics[width=\linewidth]{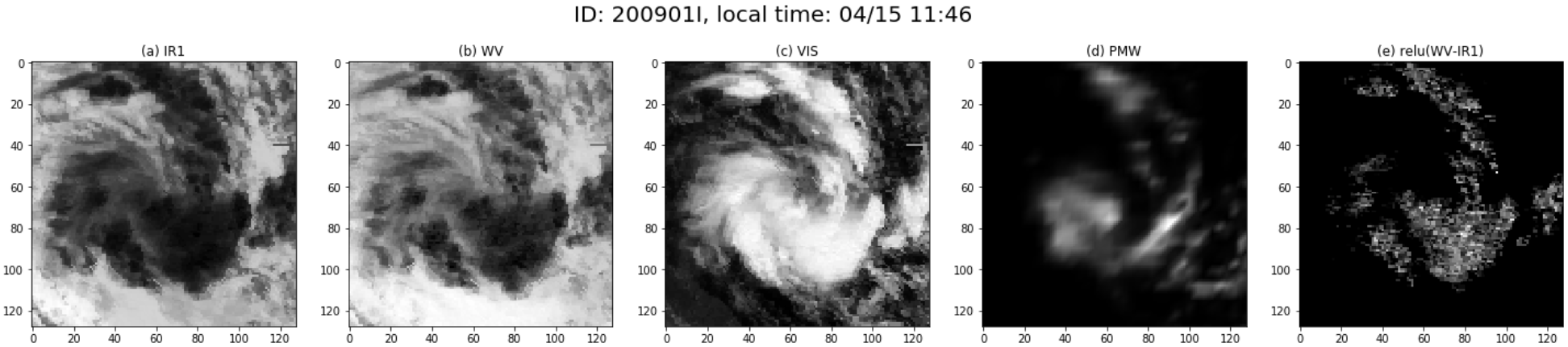}
    \caption{Channels of the TCIR dataset. (a)-(d) The four basic channels: IR1, WV, VIS, and PMW. (e) $\texttt{relu}(WV-IR1)$. By comparing (d),(e), we can find the implicit correlations between the PMW and other channels.}
	\label{fig:channels}
\end{figure}

Recently, several studies have applied convolutional neural networks (CNN) on satellite images to estimate TC intensity \cite{pradhan2018tropical, chen2018rotation, chen2019estimating, wimmers2019using}.
The previous work released a benchmark dataset "TCIR" for the TC-image-to-intensity regression task~\cite{chen2018rotation}, which consists of satellite images including four channels (\Cref{fig:channels} (a)-(d)).
The CNN-TC network\cite{chen2019estimating} \footnote{This article is published in \textit{Weather and Forecasting}, one of the best journals in the relevant research field.} utilized IR1 (Infrared) and PMW (passive microwave rain rates) channels and achieved the state-of-the-art performance with also a three-hour frequency.

Notably, the VIS (visible) channel was not utilized in the past, because it only provides meaningful cloud information during the daytime. Besides, PMW channel images can only be collected at the frequency of $\approx3$ hours while the observations from IR1 and water vapor (WV) channel  are available almost anytime.\footnote{PMW is collected by low-Earth-orbit microwave satellites. Meanwhile, IR1, WV, and VIS are collected by geostationary satellites. Normally, geostationary satellites collect data at the frequency of 15 min. In a super rapid scan mode, a geostationary satellite provides observations every 2 min.} As real-time intensity estimations are required for pragmatic disaster management, we must be able to handle temporally-heterogeneous satellite observations.

% Furthermore, systematical examining TC activity in the past few decades is of importance for understanding extreme weather events in the rapidly changing and warming climate \cite{knutson2019tropical}. It is thus needed to provide intensity estimates for TCs back to 1980-90s, during that time, only geostationary satellite observation was available \cite{knapp2011globally}. 

\begin{figure}[t]
    \centering
    \includegraphics[width=0.8\linewidth]{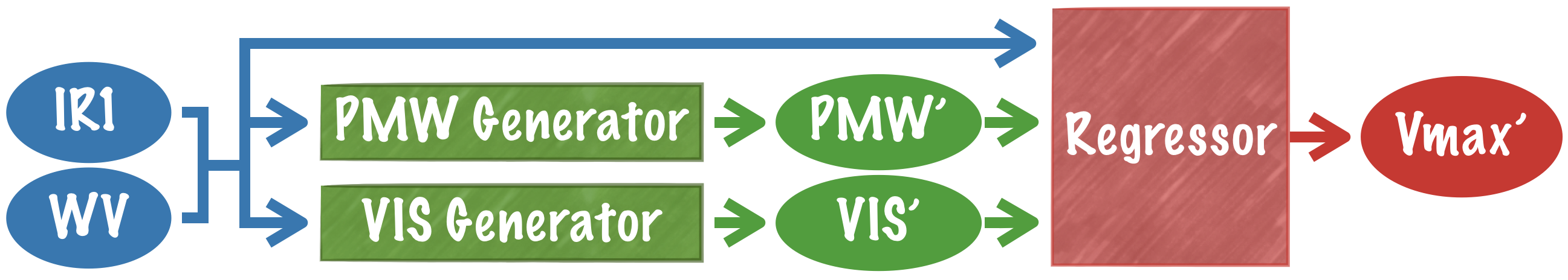}
    \caption{A schematic showing how the proposed hybrid GAN-CNN model is used for estimating TC intensity. Note that only the temporally homogeneous IR1 and WV data go into the model as the input while the GAN component generates the PMW and VIS features for the CNN regressor.}
	\label{fig:predicting_model}
\end{figure}

This work proposes a novel deep learning model combining CNN and GAN (generative adversarial network) to deal with the temporally heterogeneous datasets.
Our goal is to eliminate the dependency on PMW channel and perform the prediction with only IR1 and WV channels, so that 24/7 intensity estimations could be provided.
To keep the performance comparable to the state-of-the-art model, we also make use of the most of the information provided by good quality VIS images.
We proposed a novel 5-stage training strategy, with which two separated generators are trained for producing simulated VIS and PMW images. Afterward, generated VIS and PMW images can be used along with IR1 images to conduct the estimates. \cref{fig:predicting_model} is a schematic of our inference model.

We summarize related work about GAN in \cref{sec:related} and provide analysis of various used satellite observations in \cref{sec:data}.
\cref{sec:method} discusses the hybrid GAN-CNN architecture and the training strategy. 
Then \cref{sec:experiment} demonstrates the capability of the proposed model dealing with temporally heterogeneous datasets to provide 24/7 intensity estimates in good quality.
 \cref{sec:conclusion} is a quick recap.

\section{Background Knowledge}
\label{sec:related}

In this section, we summarize several GAN frameworks that inspire our work. In a typical \textbf{GAN} \cite{goodfellow2014generative}, there are two opposing players: a \emph{generator} and a \emph{discriminator}.
The generator is responsible for generating fake data and trying to confuse the discriminator.
A discriminator acts like an umpire, responsible for distinguishing between real and fake data. The competition between both players prompt the generator to generate fake data that is difficult to distinguish from the real data.

\textbf{CGAN} (conditional GAN) \cite{mirza2014conditional} attaches conditions to the input of the generator.
These conditions should be related to several side information provided by the image, such as the class of the object in the picture.
The generator is restricted only to generate images that meet the conditions.
Also, the specified conditions will be disclosed to the discriminator along with the generated data.
This formulation allows generators to generate images according to our needs.

\textbf{AC-GAN} (auxiliary classifier GAN) \cite{odena2017conditional} follows the steps of CGAN by setting conditions to the generator. Differently, specified conditions are not exposed to the discriminator.
Instead, the discriminator are demanded to reconstruct the side information of the images on its own.
Compared to CGAN, AC-GAN further improves the stability of training and the quality of the generated data.

In most of GAN frameworks, the discriminator only outputs a single probability to determine whether the entire image is generated or not.
In contrast, \textbf{PatchGAN} \cite{li2016precomputed} modifies its discriminator to cut the whole image into multiple small patches with overlap and discriminates them piece by piece.
This technique has been proven to be mathematically equivalent to doing blending with data after cutting into patches.
This technique is useful for data with even distribution and no distinct boundary, such as satellite images.

\textbf{U-Net} \cite{ronneberger2015u} is a type of generators which has similar structure to an auto-encoder. The input and output of U-Net are both images.
It retains the local details from the input image by skip connections, then reconstructs them in the corresponding position.
Therefore, it is widely used in situations where the input and output are pictures of the same size.

Proposed in 2017, \textbf{Pix2Pix} \cite{isola2017image} uses pictures as the condition of the generator and completes the style conversion task brilliantly.
Because the inputs of the generator are images, Pix2Pix reasonably uses U-Net as its generator. Meanwhile, the PatchGAN discriminator is used.
L1 distance between the input picture and the output picture are added to the generator loss.
This term in the loss function directly guides the generator to produce the desired image and is useful for fighting against \emph{mode collapse}.
However, this framework demands data before and after conversion to be paired.

\textbf{CycleGAN} \cite{zhu2017unpaired} focuses on training two generators (style A-> style B, style B-> style A) at the same time instead of training an one-way generator.
CycleGAN requires the input image (assuming style A) can be converted back as similar as possible after passing through both generators.
We can thus ensure that the generator retains the critical information in the original image when converting the style. 
The concept in CycleGAN is similar to an auto-encoder.
On the other hand, CycleGAN also resolves the limitation of Pix2Pix that requires the paired data.
However, due to the characteristics of TC images,
we apply Pix2Pix in our framework, but not that of the CycleGAN. The detailed reasons are described in \cref{sec:pmw}.
\section{Data Analysis and Preprocessing}
\label{sec:data}
We conducted our experiment on the benchmark TCIR dataset, which includes 4 channels: Infrared (IR1), Water Vapor (WV), Visible (VIS), and Passive Microwave (PMW). An example is shown in \cref{fig:channels} (a)-(d).
For more details about the TCIR, please refer to the previous work of \citet{chen2018rotation}.

\subsection{Analysis of the PMW Data }
\label{sec:pmw}
According to domain knowledge, PMW is positively correlated with \textit{relu(WV-IR1)}  \cite{olander2009tropical}.
Therefore, the \cref{fig:channels}(d) and \cref{fig:channels}(e) are somewhat similar to each other.
% \begin{figure}[h]
% \includegraphics[width=\linewidth]{figs/PMW_correlation.png}
% \caption{PMW and relu(WV-IR1) correlation.}
% 	\label{fig:pmw}
% \end{figure}
This evidence makes us believe that a properly trained Pix2Pix model can hopefully restore the PMW channel using IR1 and WV information.
However, since there is still a certain gap between PMW and \textit{relu (WV-IR1)}, we cannot directly use the latter one to replace the former one.

Besides, we also discovered that the conversion is uni-directional.
While we can use IR1 and WV to derive PMW, it is challenging to derive IR1 and WV from PMW.
Thus, cycleGAN is not recommend to serve as the PMW generator.

\subsection{Analysis of the VIS Data}
\label{sec:vis}
The VIS channel is the noisiest compared with other three channels.
Types of effects includes:
\begin{compactenum}
    \item VIS images are meaningless at night due to the significant decrease in light intensity.
    \item Even under the daylight, about 1/5 of the VIS are noisy or completely black.
    \item The intensity of sunlight varies at different hours throughout the day. Cloud is more obvious when the time is closer to noon.
    \item Similar to other channels, the signal may be disturbed by noises. Sometimes even half of a VIS image is black. Besides, there could be strip noise occasionally.
\end{compactenum}
By adopting the VIS generator, we want to (1) generate VIS images with IR1 and WV at all times, even during the night, (2) calibrate sunlight level among all VIS images, and (3) remove block and strip noise. For clearer example, please refer to \cref{fig:generation} in the experiment section.

\subsection{VIS Quality Control}
\label{sec:good_quality}
% Before training the VIS generator, We need to mark out a subset of VIS images which have good quality to learn from.
A subset of VIS data with good quality is needed to serve as the labeling data and facilitate the training of the VIS generator. This selection was conducted based on the calculation of the mean and standard deviation of the entire image values.

By filtering unusual values of mean, we can exclude all-black and all-white images.
On the other hand, a clear VIS image's standard deviation is likely to fall within a certain interval. After consulting human experts,
we subjectively determine the range of reasonable mean and standard deviation in which a high-quality VIS should be (1) $0.1 \ge mean \ge 0.7$ and (2) $0.1 \ge std \ge 0.31$.

Moreover, to further reduce false positives, we limit high-quality visible images to be between 07:00 and 17:00 because it is impossible to have high-quality VIS data during the nighttime, namely our third condition: (3) $7 \geq local\_time.hour \ge 17$.

As a result, a total of 12480 TC images (training data) were marked as good quality VIS, accounting for 20.9\% of the total 59837 samples in the training data.

% The distribution throughout the day is shown in \cref{fig:good_distrib}.
% Even at around noon, data with good quality in the VIS channel only accounts for about 60\%.
% This fact explains that, in the experimental results in our previous work \cite{chen2018rotation}, including VIS observation into the CNN model does not lead to better performance.

% \begin{figure}[t!]
% \includegraphics[width=\linewidth]{figs/good_quality_VIS_count.png}
% \caption{Bar charts of the sample numbers of various time in a day for all VIS data (green) and good-quality VIS (blue).}
% \label{fig:good_distrib}
% \end{figure}

\section{Proposed Method}
\label{sec:method}
% In this section, we will first introduce our hybrid GAN-CNN model framework, consist of two sets of GANs and a CNN regressor.
% To make our two separated GANs trained properly and cooperating well, a novel 5-stage training strategy is proposed in the second half of this section.

\begin{figure}[t]
\centering
\begin{minipage}{.48\textwidth}
  \centering
  \includegraphics[width=0.99\linewidth]{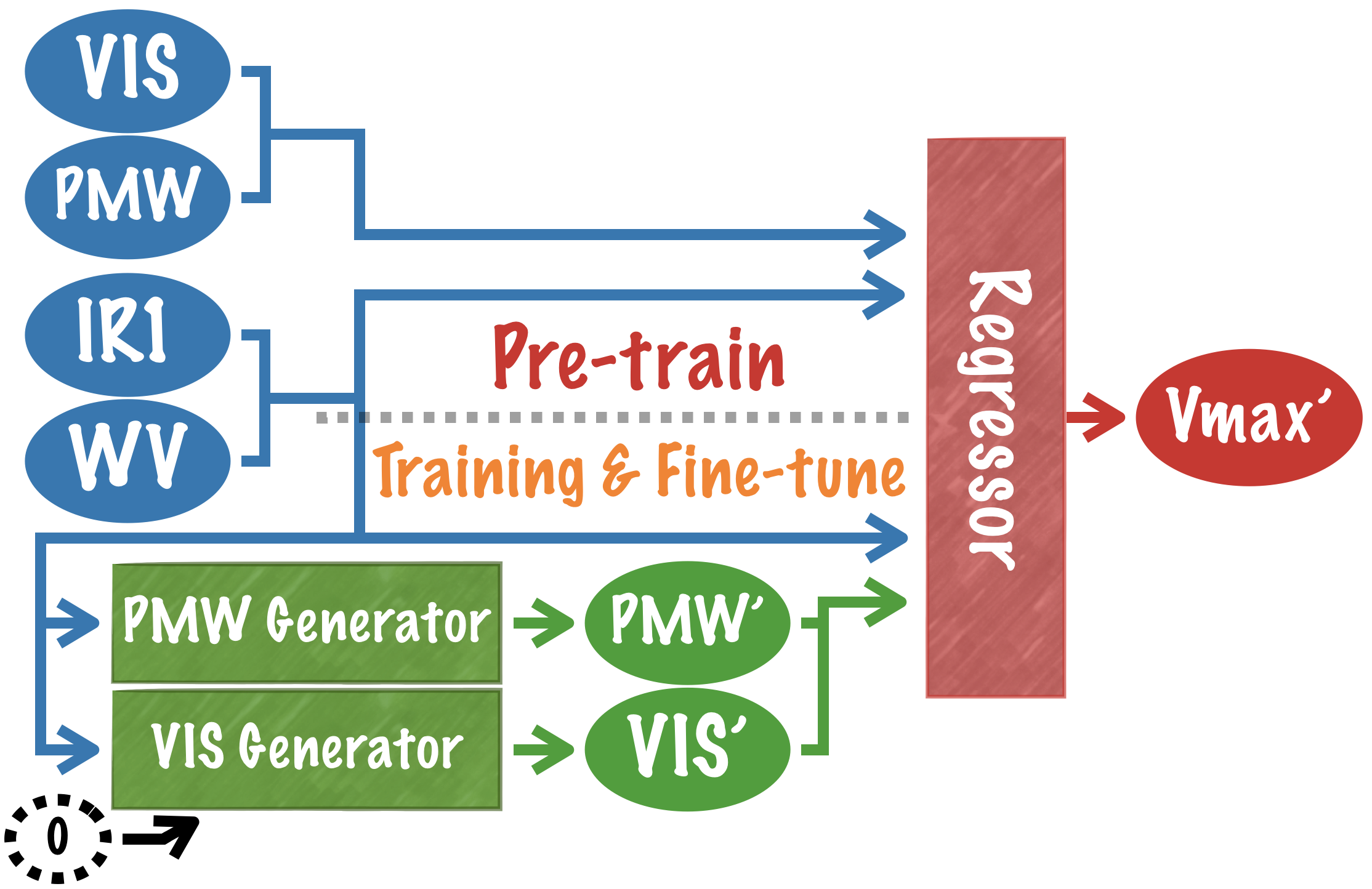}
  \caption{Framework of hybrid GAN-CNN.}
  \label{fig:compound_model}
\end{minipage}\hspace{2mm}
\begin{minipage}{.48\textwidth}
  \centering
  \includegraphics[width=0.99\linewidth]{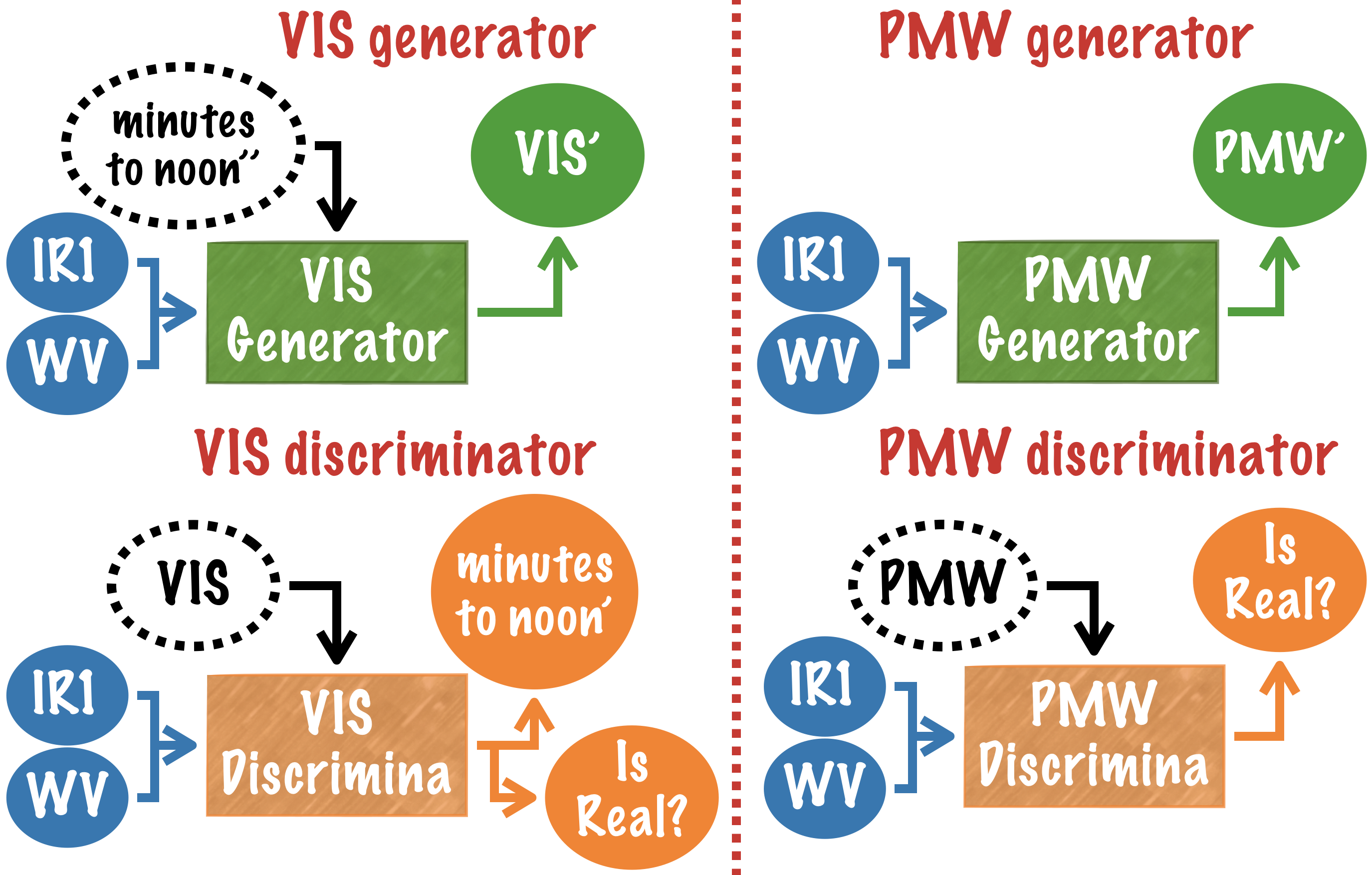}
  \caption{Illustration of two sets of GANs.}
  \label{fig:gan_structure}
\end{minipage}
\vspace{-2mm}
\end{figure}

To eliminate the dependence on the usage of PMW and solve problems caused by severe noise in the VIS,
we demand GANs that can stably generate VIS and PMW channels only by the two temporally homogeneous channels: IR1 and WV (\cref{fig:compound_model}). 
In the dataset, every set of TC images have PMW observation, while only 20\% have good-quality VIS.
To use available information optimally, generators for PMW and VIS are trained separately (\cref{fig:gan_structure}).

%Similar to Pix2Pix \cite{isola2017image},
We use an adjusted \textbf{U-Net} for the \textbf{generator}.
Thanks to the skip connections in U-Net, local details in the source image can be preserved.
% Because the four channels are aligned with respect to the TC center (i.e.,  the center point of the image), it is suitable for the U-Net architecture.
Regarding the \textbf{discriminator}, we use the idea of \textbf{PatchGAN}.
Satellite observations are continuous, without concept of objects, foreground, background, and boundaries.
Radically speaking, any segmentation of a real TC image can be determined normal because there is no object to be dissected.
A PatchGAN divides images into multiple small areas before determining whether it is real, which is suitable for our data and makes training more stable.

% \begin{figure}[t!]
% \centering
% \includegraphics[width=.8\linewidth]{figs/two_set_of_GAN.png}
% \caption{Illustration of the framework for two sets of GANs.}
% 	\label{fig:gan_structure}
% \end{figure}

% \begin{figure}[t!]
% \centering
% \includegraphics[width=.8\linewidth]{figs/compound_model_illustration.png}
% \vspace{1mm}
% \caption{Framework of our hybrid GAN-CNN model.}
% 	\label{fig:compound_model}
% \end{figure}

\subsection{Training Objective}
\label{sec:loss}
There are three components in this framework: a generator, a discriminator, and a regressor.
The loss functions are formulated as:
\begin{equation}
\begin{aligned}
\mathcal{L}_{D_{target}} =& \; {l}_{disc} + \gamma \times {l}_{m2n} \times [[target = vis]] \\
\mathcal{L}_{G_{target}} =& \; {l}_{gen} + \alpha \times {l}_{L2} + \beta \times {l}_{regr} + \gamma \times {l}_{m2n} \times [[target = vis]] \\
\mathcal{L}_{R} =& \; {l}_{regr}, \;\;\;\;\;\;
target \in \; \{vis, pmw\}
\end{aligned}
\label{loss:overview}
\end{equation}
% Each of the above loss terms is detailed below.
$\alpha, \beta, \gamma$ used in the experiments are listed in \cref{appendix:hyper}.

\paragraph{Discriminator loss ($\mathcal{L}_D$)}
The goal of the discriminator is to correctly identify the target image as real observation or a fake generated by the generator:
\begin{equation}
\begin{aligned}
{l}_{disc}(G, D) = \mathop{\mathbb{E}}_{ir1, wv, target}[\log D_{disc}(ir1, wv, &target) + 1 - \log D_{disc}(ir1, wv, target')], \\
target' =  G_{target}(ir1, wv),& \; target \in \{vis, pmw\}
\end{aligned}
\label{loss:disc}
\end{equation}
This concept is the same as that in the Pix2Pix.

\paragraph{Generator loss ($\mathcal{L}_G$): }
The loss contains the following requirements associated with our loss function.
First, the goal of the generator is to confuse the discriminator:
\begin{equation}
\begin{aligned}
{l}_{gen}(G, D) = \mathop{\mathbb{E}}_{ir1, wv, target} &\log D_{disc}(ir1, wv, target'),\\
target' = G_{target}(ir1, &wv), \; target \in \{vis, pmw\}
\end{aligned}
\label{loss:gen}
\end{equation}
This is equivalent to the loss in the patchGAN.

Second, the generated VIS/PMW must be similar to the input VIS/PMW:
\begin{equation}
\begin{aligned}
{l}_{L2}(G) = \mathop{\mathbb{E}}_{ir1, wv, target} &||target - target'||_2, \\
target' = G_{target}(ir1, &wv), \; target \in \{vis, pmw\}
\end{aligned}
\label{loss:L2}
\end{equation}
This concept is borrowed from the Pix2Pix. Note, however, that the L2 distance (MSE) is used instead of the L1 distance (MAE) in our model because L2 distance encourages more blurring.
Usually, we want to generate images with clear lines and apparent boundaries.
However, satellite images have no concept of boundaries, where they are smoother than ordinary pictures, such as dogs, cats, and cars. 
Therefore, we modify the Pix2Pix architecture based on our need.
A comparison of using L1 distance and L2 distance will be shown later in \cref{sec:l1_l2}.

Besides, we have added the following two innovative designs to our GANs, specializing the GANs to complete tasks appropriately.
The details are described in the following two sub-sections.

\paragraph{Auxiliary time loss ($l_{m2n}$): }
As mentioned in \cref{sec:vis},
we expect the VIS generator to adjust all generated VIS images to the sunlight level at noon.
To achieve this goal, we need to first calculate m2n (minutes to noon) for each VIS image:
$m2n = | 60 \times hour + minute-60 \times 12 |$.
We take m2n as an additional condition and apply the concept of the AC-GAN to our VIS generator and discriminator.
Then loss function therefore has an extra term.\\
Discriminator:
\begin{equation}
\begin{aligned}
{l}_{m2n}(D) = \mathop{\mathbb{E}}_{vis, m2n} ||m2n-D_{m2n}(vis)||_2
\end{aligned}
\label{loss:m2n_d}
\end{equation}
Generator:
\begin{equation}
\begin{aligned}
{l}_{m2n}(G, D) = \mathop{\mathbb{E}}_{ir1, wv} ||m2n'-&D_{m2n}(vis')||_2, \\
vis' = G_{vis}(ir1, wv, m2n'),& \; 0 \leq m2n' \leq 300
\end{aligned}
\label{loss:m2n_g}
\end{equation}

VIS images with good quality were obtained during daytime, specifically from 7:00 to 16:59 (see \cref{sec:good_quality}), so the largest value of minutes to noon is 300.
In training, randomly generated floating-point numbers between [0, 300] are provided to the generator as condition.

\paragraph{Regressor loss ($\mathcal{L}_R$): }
Since our ultimate goal is to use the generated VIS / PMW channels as the regressor's inputs, the generator is requested to create useful features that can facilitate estimations.

The generator is first asked to generate PMW and VIS given the condition $m2n = 0$.
The obtained results are provided to the regressor, along with IR1 and WV.
Finally we can obtain our intensity estimation using generated VIS / PMW (e.g. lower part of \cref{fig:compound_model}).
The precision of the prediction is also added to one of the terms of the generator loss:
\begin{equation}
\begin{aligned}
 {l}_{regr}(G_{vis}, G_{pmw}, R) &= \mathop{\mathbb{E}}_{ir1, wv, vmax}||vmax-vmax'||_2, \\
vmax' = R (ir1, wv, vis', pmw'), \;&
vis' = G_{vis}(ir1, wv, 0), \;
pmw' = G_{pmw}(ir1, wv)\\
\end{aligned}
\label{loss:regr}
\end{equation}
The term $vmax$ stands for the maximum wind velocity, the definition of TC intensity.

\subsection{Strategy of Three-Stage Training}
\label{sec:3stage}

% \begin{figure}[t!]
% \centering
% \includegraphics[width=.8\linewidth]{figs/compound_model_illustration.png}
% \vspace{1mm}
% \caption{Framework of our hybrid GAN-CNN model.}
% 	\label{fig:compound_model}
% \end{figure}

\begin{figure}[t]
\centering
\begin{minipage}{.28\textwidth}
    \centering
    \includegraphics[width=.99\linewidth]{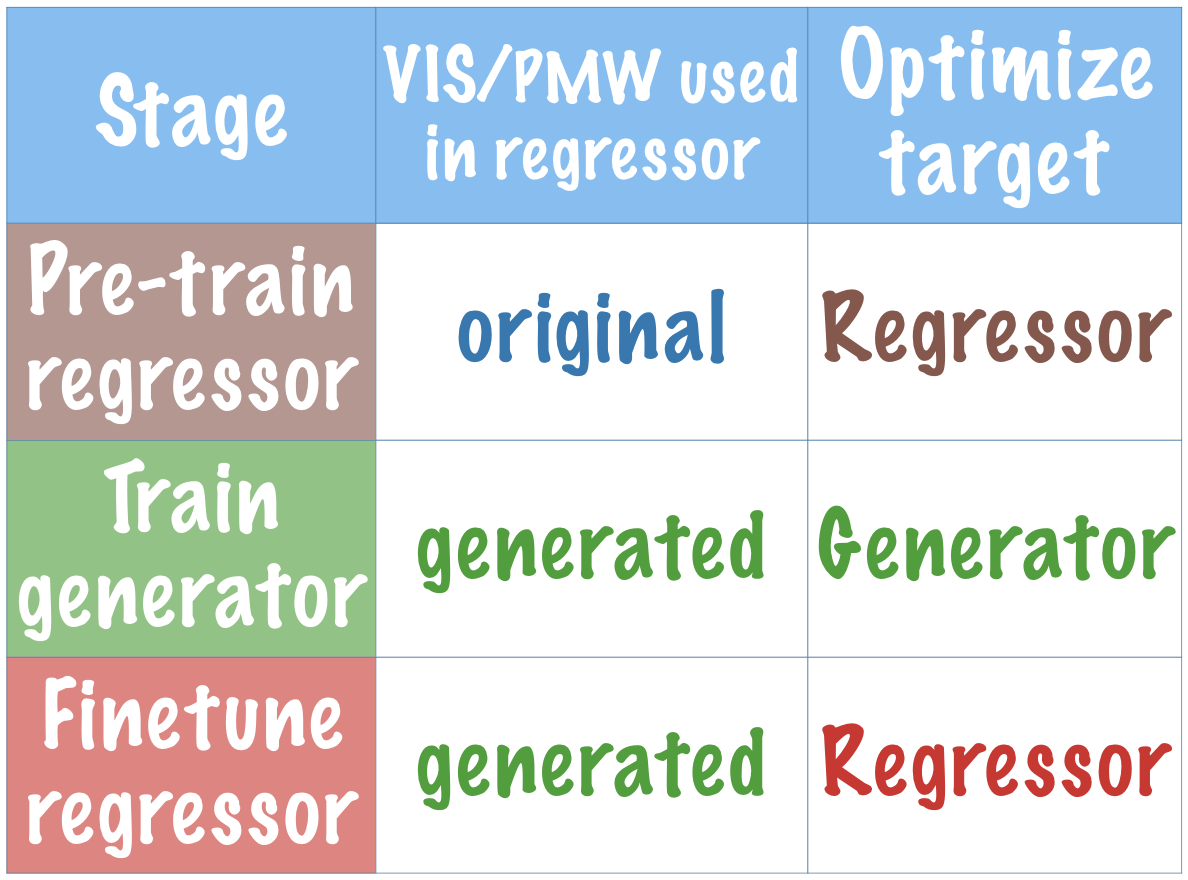}
    %\captionsetup{width=\linewidth}
    \caption{3-stage training.}
    \label{fig:3stage}
\end{minipage}\hspace{2mm}
\begin{minipage}{.68\textwidth}
    \centering
    \includegraphics[width=.99\linewidth]{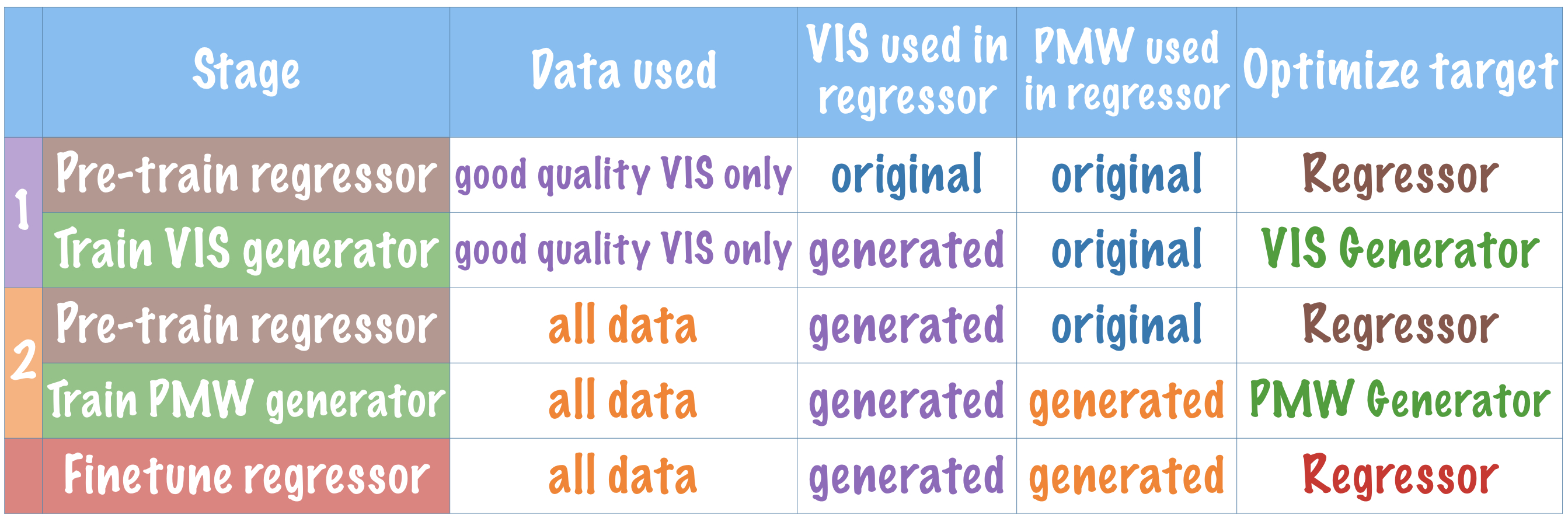}
    %\captionsetup{width=.8\linewidth}
    \caption{The steps of the extended 5-stage training.}
    \label{fig:5stage}
\end{minipage}
\end{figure}

To fairly calculate ${l}_{regr}$, we need to pre-train the regressor.
Therefore, a novel three-stage training illustrated in \cref{fig:compound_model} and \cref{fig:3stage} is proposed.

% \begin{figure}[t!]
% \centering
% \includegraphics[width=\linewidth]{figs/three_stage_training.png}
% \vspace{1mm}
% \caption{The steps of the 3-stage training.}
% 	\label{fig:3stage}
% \end{figure}

% \begin{figure}[t!]
% \centering
% \includegraphics[width=\linewidth]{figs/five_stage_training.png}
% \caption{The steps of the extended 5-stage training.}
% 	\label{fig:5stage}
% \end{figure}

\begin{compactenum}
    \item \textbf{Pre-train regressor}: Pre-train the regressor with original VIS and PMW along with other regressor inputs and thereby let the regressor learns how to extract essential features from real VIS / PMW in advance.
    \item \textbf{Train generator}: The generator and discriminator are optimized together.
    The loss mentioned in \cref{loss:regr} ensures that significant feature regressor is using will be generated.
    \item \textbf{Fine-tune regressor}: With well-trained generator, we use the generated VIS / PMW features to fine-tune the regressor.
\end{compactenum}

\subsection{Strategy of Five-Stage Training}
\label{sec:5stage}

Good results are achieved through the above procedures of three-stage training. To be even better, VIS generator should be trained with only data with good quality VIS images.
Meanwhile, the training process become more stable when 2 generators are only optimized when another one is fixed.

Therefore, based on the three-stage training described above,
a more detailed five-stage training process illustrated in \cref{fig:5stage} is proposed.
In the five-stage training, the first two stages of three-stage training are repeated for two loops:
\begin{compactenum}
\item Loop \textbf{1}:
Only data with good quality VIS images are used in this loop. PMW generator is fixed while we focus on optimizing VIS generator. When calculating loss in \cref{loss:regr} we use the original PMW instead of a generated one.
\item Loop \textbf{2}: All data are used. VIS generator is fixed while we pay attention to PMW generator.
\item After training both generators, regressor get fine-tuned using the generated PMW' and VIS'.
\end{compactenum}
By applying five-stage training, we obtain the final operational predicting model (\cref{fig:predicting_model}).

\section{Experiments and Analysis}
\label{sec:experiment}
%In the following section, we provide details about hyper-parameters and model parameters.
In the following section, two techniques proposed in the previous work will be explained briefly, including \textbf{auxiliary features} and \textbf{rotation-blending}. Please refer to \citet{chen2019estimating} for more details.
Next, we will compare our performance with related works which also focus on estimating TC intensity, including both operational meteorological models and deep learning techniques.
Finally, we qualitatively analyze the quality of the proposed model.
\emph{Detailed model structures and hyper-parameters are disclosed in \cref{appendix:structure} and \cref{appendix:hyper} respectively.}

In the experiments, we split the dataset into three parts: (1) Training data: TCs during 2000-2014, (2) Validation data: TCs during 2015 and 2016, (3) Testing data: TCs during 2017.

\paragraph{Auxiliary Features:}
In addition to the output from convolution layers, additional features are passed into the regressor.
The auxiliary features are demonstrated to be helpful in improving the precision of estimation \cite{chen2019estimating}. 
These features provide clues such as (1) day of year: stand for seasonal information, (2) local time, and the most influential one: (3) One-hot encoded region codes: region codes is in \{\textit{WPAC, EPAC, CPAC, ATLN, IO, SH}\}, representing 6 different basins.

\paragraph{Rotation Blending: }
Considering the nature of TCs as a rotating weather system, TC data is rotation invariant.
That is, rotations with respect to the center usually do not affect the estimation of the TC intensity.
\cite{chen2018rotation} demonstrated that the idea of using rotation for augmentation leads to a significant improvement in performance.

During the training phase, each image will be randomly rotated by any degree before feeding into our model.
When it comes to inference, images will be rotated by evenly distributed ten angles ranged from 0 to 360 to collect 10 different estimations. Afterward, these intensity estimations are blended to obtain the final estimate.

\subsection{Intensity Estimation Performance}
\label{sec:R_result}
The main task of this work is to accurately estimate the TC intensity, which is the output of our model.
The unit of TC intensity is knot (kt) defined as its maximum wind speed (Vmax). 
The value of Vmax is usually ranged in [30, 180], and TCs with a Vmax larger than 96 kt are considered as intense TCs.

\cref{tab:performance} compares our performance to other works.
ADT (Advanced Dvorak Technique) \cite{olander2007advanced} is a common used method to estimate TC intensity, which extract features from IR1 images before applying linear regression. 
SATCON \cite{velden2014update} is the the state-of-the-art model used by meteorologists in operational forecast. It highly rely on observations from low-Earth-orbit satellites.

The performance of our proposed model is comparable to the state-of-the-art model in both deep learning and meteorology, while our model can provide much more timely estimations.

\begin{figure}[t]
\centering
\begin{minipage}{.48\textwidth}
    \centering
    \includegraphics[width=.99\linewidth]{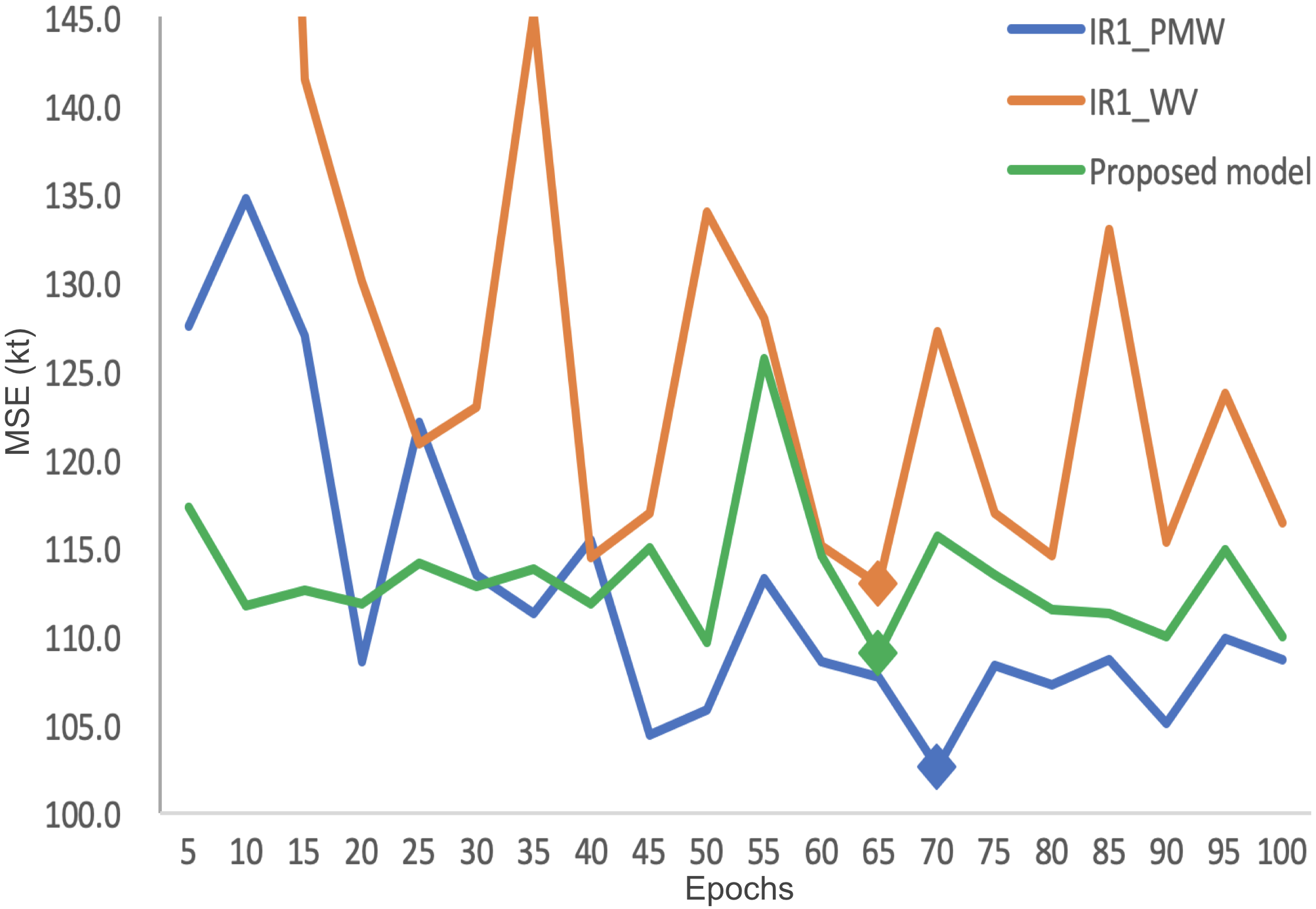}
%    \captionsetup{width=.8\linewidth}
    \caption{Learning curves in MSEs for models with different channel combinations as the input.}
    \label{fig:gan_effect}
\end{minipage}\hspace{3mm}
\begin{minipage}{.48\textwidth}
    \centering
    \includegraphics[width=.99\linewidth]{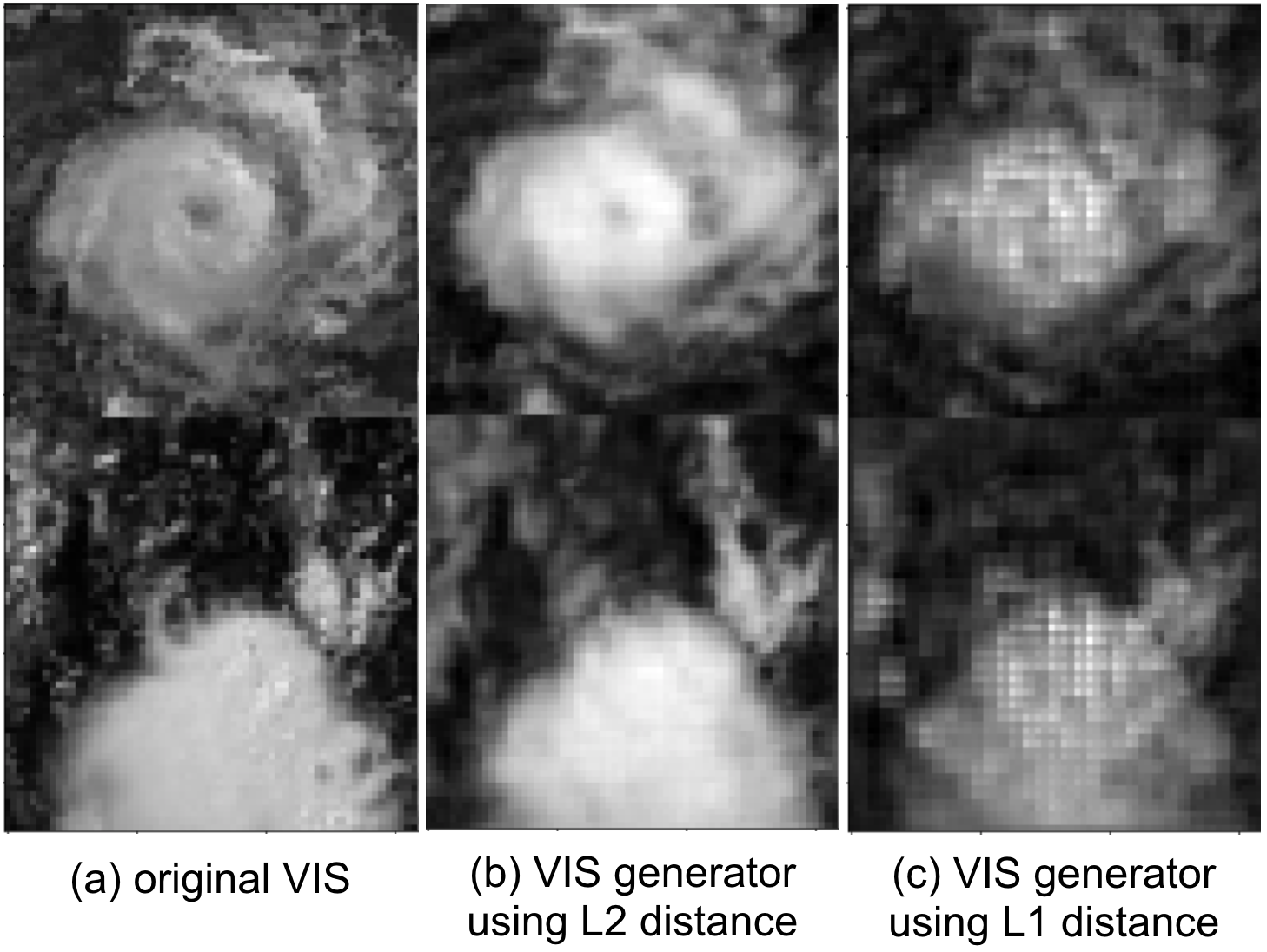}
%    \captionsetup{width=\linewidth}
    \caption{VIS examples generated by generators using L1 and L2 distance for ${l}_{regr}$, described in \cref{loss:L2}. Validation data is used for generating VIS in this figure.}
    \label{fig:l1_l2}
\end{minipage}
\vspace{-3mm}
\end{figure}

In \cref{fig:gan_effect}, we compare the validation MSE score over the first 100 epoch of training.
The blue line represents the state-of-the-art model, which provides intensity estimates every 3 hours.
The orange and green line shares same inputs, IR1 and WV, which is available every 15 min.
The former is the model that directly uses them for estimations while the latter is our proposed model.
In contrast, our adequately trained GAN model helps us further improve the performance of intensity estimation, bringing it closer to the state-of-the-art model. Most importantly, our proposed model can provide intensity estimates every 15 min.

\begin{table}[]
\centering
\caption{The comparison between RMSEs of our proposed models and state-of-the-art models.}
\label{tab:performance}
\begin{tabular}{lllllll}
\hline
 & \multicolumn{2}{l}{no smoothing} & \multicolumn{2}{l}{w/ smoothing\tablefootnote{Simple smoothing techniques are applied here to obtain a boost in estimation precision \cite{chen2018rotation, chen2019estimating}.}} & \multirow{2}{*}{input} & \multirow{2}{*}{frequency} \\ \cline{2-5}
 & valid & test & valid & test &  &  \\ \hline
ADT \cite{olander2007advanced} & \multicolumn{4}{c}{12.65} & IR1 & 30 min \tablefootnote{They are currently providing estimations every 30 mins. But it could be <=15 min as well.} \\
SATCON \cite{velden2014update} & \multicolumn{4}{c}{8.59} & IR1, \textit{PMW}\tablefootnote{SATCON depends on low-Earth-orbit satellites observations, which is somehow similar to the PMW.} & $\approx3$H \\ \hline
CNN-TC \cite{chen2019estimating} & 10.38 & -- & 8.74 & 8.39 & IR1, PMW & $\approx3$H \\ \hline
\textit{CNN-TC \tablefootnote{Our reproduction of CNN-TC. We add additional batch normalization layers in our reproduced CNN-TC, which leads to a minor improvement. The modified structure is disclosed in \cref{tab:regressor}.}} & 10.13 & 10.13 & 8.62 & 8.89 & IR1, PMW & $\approx3$H \\
Proposed model & 10.43 & 10.19 & 9.01 & 9.33 & IR1, WV & \textbf{<= 15 min} \\ \hline
\end{tabular}
\end{table}

% \begin{table}[]
% \centering
% \caption{The comparison between RMSEs of our proposed models and state-of-the-art models.}
% \label{tab:performance}
% \begin{tabular}{@{}lllllll@{}}
% \toprule
%  & \multicolumn{2}{l}{no smoothing} & \multicolumn{2}{l}{w/ smoothing\tablefootnote{Simple smoothing techniques are applied here to obtain a boost in estimation precision \cite{chen2018rotation, chen2019estimating}.}} & \multirow{2}{*}{input} & \multirow{2}{*}{frequency} \\ \cmidrule(lr){2-5}
%  & valid & test & valid & test &  &  \\ \midrule
% ADT \cite{olander2007advanced} & \multicolumn{4}{c}{12.65} & IR1 & 30 min \tablefootnote{They are currently providing estimations every 30 mins. But it could be <=15 min as well.} \\
% SATCON \cite{velden2014update} & \multicolumn{4}{c}{8.59} & IR1, \textit{PMW}\tablefootnote{SATCON depends on low-Earth-orbit satellites observations, which is somehow similar to the PMW.} & $\approx3$H \\ \midrule
% CNN-TC \cite{chen2019estimating} & 10.38 & -- & 8.74 & 8.39 & IR1, PMW & $\approx3$H \\ \midrule
% \textit{CNN-TC \tablefootnote{Our reproduction of CNN-TC. We add additional batch normalization layers in our reproduced CNN-TC, which leads to a minor improvement. The modified structure is disclosed in \cref{tab:regressor}.}} & 10.13 & 10.13 & 8.62 & 8.89 & IR1, PMW & $\approx3$H \\
% Proposed model & 10.43 & 10.19 & 9.01 & 9.33 & IR1, WV & \textbf{<= 15 min} \\ \bottomrule
% \end{tabular}
% \end{table}

\subsection{Effectiveness of the L2 Distance}
\label{sec:l1_l2}
L2 distance is chosen instead of L1 distance in \cref{loss:L2}, which is different from a ordinary Pix2Pix framework.
Comparing to commonly seen pictures, satellite observations are continuous and have smoother boundaries.
As described in \cref{sec:loss}, using L2 distance encourages more blurring.
In \cref{fig:l1_l2}, we take VIS channels as examples, compare the generated results from models using L1 distance and L2 distance.
As shown, the model using L1 distance generates images less smoothly.

\subsection{Qualitative Study}
\label{sec:G_result}
\begin{figure}[t!]
\centering
\includegraphics[width=\linewidth]{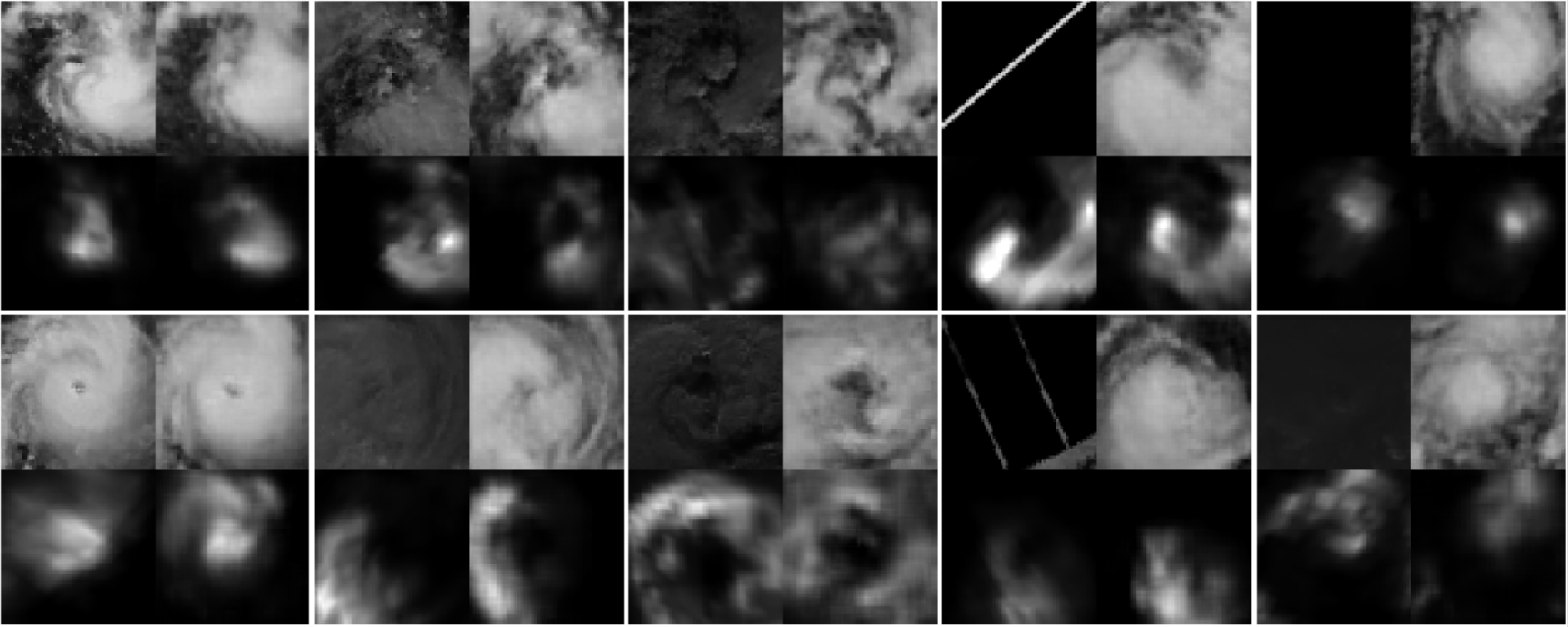}
\caption{The generated VIS and PMW data. For each block, there are original VIS (upper left), generated VIS (upper right), original PMW (lower left), and generated PMW (lower right). Validation data is used for generating this figure.}
	\label{fig:generation}
\end{figure}

\cref{fig:generation} shows the generated images from the proposed model.
Compared with the original VIS images, the generated VIS image is slightly blurred, and the eye is not as clear as the original one.
Nevertheless, these images can be generated stably anytime, with the removal of most noises, and with adjustment of sunlight intensity.

Interestingly, most of the generated PMW seems to be slightly rotated. This is presumably because observations can only be obtained when the low-Earth-orbit satellites pass across the TCs. Therefore, the time PMW channel being collected could be slightly misaligned with other channels.

\section{Conclusion}
\label{sec:conclusion}
This paper focuses on improving the utility of deep learning in TC intensity estimation for practical scenarios.
A modified Pix2Pix GAN framework is presented to better fit in our unusual TC data that is temporally-heterogeneous, and we combine the GAN with a CNN regressor, eventually becoming our proposed hybrid GAN-CNN model.
By properly dealing with temporally-heterogeneous data, our proposed model not only achieves comparable performance with the state-of-the-art model in estimating precision but also provides the estimates much more timely, which leads to better TC warning operations and could eventually save more lives.

% In the future, we can further combine the state-of-the-art model and our proposed model, one with better precision and another with a higher frequency of estimation along with comparable accuracy.
Moreover, being capable of estimating TC intensity by only IR1 and WV channels, the new model enables us to reanalyze the intensity of TCs further back to the time only simple IR1 and WV images were available (1980s) instead of starting from 2000s, providing the potential of a breakthrough in the research about climate change and global warming from the aspect of TC activities.

%in the research about TC activities in the changing and warming climate.

%This novel design provides a high practical value, leads to a more timely TC warning, and has the potential to re-analyze global TC intensity back to the time only simple IR1 and WV images were available to understand the trend of TCs in the changing and warming climate.
\section*{Broader Impact}

Timely warnings are critical to a better disaster warning/management system.
From Hurricane Katrina to Hurricane Harvey and even to recent COVID-19,
we should learn that early awareness of these disaster could result in saving many lives.

Take Hurricane Harvey as example.
Back to August 25, 2017, Hurricane Harvey was estimated as a category 2 hurricane during 02:00 to 14:00.
After it became a category 3 hurricane at 17:00,
all in a sudden, it grew into a category 4 hurricane only 3 hours later at 20:00. Eventually, Hurricane Harvey led to the record-breaking damage in the U.S.
To timely monitor this rapid intensification is challenging for the conventional TC intensity estimation techniques, which provides estimates only once per 3 hours.
Imagine that if there's a timely estimation system which can provide intensity estimates every 15 minutes.
The insane increasing trend could be detected earlier, precious time could be bought, people could get better prepared, and lives could be saved.

On the downside, a significant pitfall of applying machine learning techniques in disaster management is that models are naturally conservative when facing extreme cases.
For example, TC intensity ranges from 20 to 180 kt, while about 70\% of the value distributed within 35 kt to 64 kt, which makes it almost unavoidable for machine learning models to under-estimate unprecedented extreme values.
Therefore, an auxiliary warning system should exist no matter how precise we humans can be with our machine learning techniques.
After all, disaster management is a cost-sensitive scenario in which we should always keep in mind that while false positive is annoying, false negative is deadly.

% \begin{ack}
% Use unnumbered first level headings for the acknowledgments. All acknowledgments
% More information about this disclosure can be found at: \url{https://neurips.cc/Conferences/2020/PaperInformation/FundingDisclosure}.

% Do {\bf not} include this section in the anonymized submission, only in the final paper. You can use the \texttt{ack} environment provided in the style file to autmoatically hide this section in the anonymized submission.
% \end{ack}

%\input{template}

\bibliographystyle{unsrtnat}
\bibliography{TCIR-GAN}

\clearpage
\appendix

\section{Model Structure}
\label{appendix:structure}
The model structures for the generator, the discriminator, and the regressor are detailed in Table~\ref{tab:generator}, Table~\ref{tab:discriminator}, and Table~\ref{tab:regressor} respectively.

\begin{table}[ht]
\centering
\caption{Model structure of the generator.}
\label{tab:generator}
\begin{tabular}{ccccccccc}
\hline
\bf i  & \bf Oper   & \bf skip & \bf Ker. & \bf Stri. & \bf Dim. & \bf BN & \bf Drop  & \bf activ. \\ \hline
0  & conv        & to 11 & 4x4  & 2x2   & 32   & N  & -   & leaky  \\
1  & conv        & to 10 & 4x4  & 2x2   & 64   & Y  & -   & leaky  \\
2  & conv        & to 9 & 4x4  & 2x2   & 128  & Y  & -   & leaky  \\
3  & conv        & to 8 & 4x4  & 2x2   & 256  & Y  & -   & leaky  \\
4  & conv        & to 7 & 4x4  & 2x2   & 256  & Y  & -   & leaky  \\
5  & conv        & -  & 4x4  & 2x2   & 256  & Y  & -   & leaky  \\
6  & trans\_conv & -  & 4x4  & 2x2   & 256  & Y  & 0.5   & relu   \\
7  & trans\_conv & -  & 4x4  & 2x2   & 256  & Y  & 0.5   & relu   \\
8  & trans\_conv & -  & 4x4  & 2x2   & 256  & Y  & -   & relu   \\
9  & trans\_conv & -  & 4x4  & 2x2   & 128  & Y  & -   & relu   \\
10 & trans\_conv & -  & 4x4  & 2x2   & 64   & Y  & -   & relu   \\
11 & trans\_conv & -  & 4x4  & 2x2   & 32   & Y  & -   & relu   \\
12 & conv        & -  & 4x4  & 1x1   & 1    & N  & -   & relu   \\ \hline
\end{tabular}
\end{table}

\begin{table}[ht]
\centering
\caption{Model structure of the discriminator; the input of discriminator layers is directly passed from shared layers.}
\label{tab:discriminator}
\begin{tabular}{lcccccc}
\hline
\bf Usage                                                                    & \bf Operation  & \bf Ker. & \bf Stri. & \bf Dim. & \bf BN & \bf activ. \\ \hline
\multirow{3}{*}{\begin{tabular}[c]{@{}l@{}}Shared\\layers\end{tabular}} & conv   & 4x4  & 2x2   & 32   & N  & leaky  \\
                                                                         & conv   & 4x4  & 2x2   & 64   & Y  & leaky  \\
                                                                         & conv   & 4x4  & 2x2   & 128  & Y  & leaky  \\ \hline
\multirow{4}{*}{\begin{tabular}[c]{@{}l@{}}Predict\\m2n\end{tabular}}   & conv   & 4x4  & 2x2   & 128  & Y  & leaky  \\
                                                                         & conv   & 4x4  & 2x2   & 256  & Y  & leaky  \\
                                                                         & linear & -  & -   & 128  & Y  & relu   \\
                                                                         & linear & -  & -   & 1    & N  & -    \\ \hline
\multirow{2}{*}{\begin{tabular}[c]{@{}l@{}}Discriminator\\layers\end{tabular}}                                            & conv   & 4x4  & 1x1   & 256  & Y  & relu   \\
                                                                         & conv   & 4x4  & 1x1   & 1    & N  & -    \\ \hline
\end{tabular}
\end{table}

\begin{table}[ht]
\centering
\caption{Model structure of the regressor. The first batch normalization layer right serves as z-score normalization. After the convolution layers, 10 dimension features are passed into linear layers along with the convolution layers' output.}
\label{tab:regressor}
\begin{tabular}{lccccc}
\hline
\bf Operation    & \bf Kernel    & \bf Strides   & \bf Dim.   & \bf BN   & \bf activ.   \\ \hline
BN        & -       & -       & -    & Y    & -     \\
conv         & 4x4       & 2x2       & 16     & Y    & relu     \\
conv         & 3x3       & 2x2       & 32     & Y    & relu     \\
conv         & 3x3       & 2x2       & 64     & Y    & relu     \\
conv         & 3x3       & 2x2       & 128    & Y    & relu     \\ \hline
\multicolumn{6}{c}{concatenate 10 additional features} \\ \hline
linear       & -       & -       & 256    & Y    & relu     \\
linear       & -       & -       & 64     & Y    & relu     \\
linear       & -       & -       & 1      & N    & -      \\ \hline
\end{tabular}
\end{table}

\section{Hyperparameters}
\label{appendix:hyper}
The hyperparameters used in the proposed model are shown in Table~\ref{tab:hyperparameters} for reproducibility.

\begin{table}[ht]
\centering
\caption{Hyper-parameters used in the 5-stage training.}
\label{tab:hyperparameters}
\begin{tabular}{lcccc}
\hline
\bf Stage               & $\alpha$(L2) & $\beta$(regr) & $\gamma$(m2n) & Max epochs \\ \hline
Pre-training Regressor & -       & -             & -        & 70         \\
VIS Generator       & 1000      & 0.0001          & 0.002      & 500        \\
Pre-training Regressor & -       & -             & -        & 100        \\
PMW Generator       & 10        & 0.001           & -        & 200        \\
Fine-tune regressor & -       & -             & -        & 300       \\ \hline
\end{tabular}
\end{table}

\end{document}